\documentclass{article}
\usepackage{graphicx} 
\usepackage{amsmath,amssymb,amsfonts,amsthm}
\usepackage{tikz-cd}
\usepackage{nicefrac}
\usepackage{csquotes}
\usepackage[numbers]{natbib}
\usepackage{booktabs}
\usepackage{parskip}
\usepackage{acronym}
\usepackage{placeins}
\usepackage{bbm}
\usepackage{listings}\usepackage[a4paper, total={6in, 8in}]{geometry}
\usepackage{lineno}
\usepackage{url}
\usepackage{authblk}

\lstdefinestyle{myListingStyle} 
    {
        basicstyle = \small\ttfamily,
        breaklines = true,
    }

\DeclareMathOperator*{\argmin}{argmin}

\tikzcdset{row sep/normal=6em}
\tikzcdset{column sep/normal=6em}

\title{Local Compositional Complexity: How to Detect a Human-readable Messsage}
\author{Louis Mahon}
\affil{School of Informatics, University of Edinburgh \\ lmahonology@gmail.com}
\date{}

\begin{document}
\acrodef{at}[AT]{assembly theory}
\acrodef{gmm}[GMM]{Gaussian mixture model}
\acrodef{lcc}[LCC]{local compositional complexity}

\maketitle

\begin{abstract}
Data complexity is an important concept in the natural sciences and related areas, but lacks a rigorous and computable definition. This paper focusses on a particular sense of complexity that is high if the data is structured in a way that could serve to communicate a message. In this sense, human speech, written language, drawings, diagrams and photographs are high complexity, whereas data that is close to uniform throughout or populated by random values is low complexity. I describe a general framework for measuring data complexity based on dividing the shortest description of the data into a structured and an unstructured portion, and taking the size of the former as the complexity score. I outline an application of this framework in statistical mechanics that may allow a more objective characterisation of the macrostate and entropy of a physical system. Then, I derive a more precise and computable definition geared towards human communication, by proposing local compositionality as an appropriate specific structure. Experimental evaluation shows that this method can distinguish meaningful signals from noise or repetitive signals in auditory, visual and text domains, and could potentially help determine whether an extra-terrestrial signal contained a message. Code to compute LCC score for image, audio and text is available at \url{https://github.com/Lou1sM/LCCScore}.
\end{abstract}

\section{Introduction} \label{sec:intro}

Complexity as a concept arises in various forms in various fields of study. The domain we are interested in here consists of static pieces of data that do not evolve and change over time, meaning that some characterisations of complexity do not apply, such as that from complex systems theory, which involves the interaction of parts and behaviour prediction \cite{bar2002general}. Common approaches to the task we are interested in, namely quantifying complexity of static data, include Shannon entropy and Kolmogorov complexity from algorithmic information theory. These methods quantify complexity as the amount of information/length of an algorithm required to represent the data perfectly. A significant drawback of these measures, with respect to the present task, is that they fail to identify random data as low complexity, and in fact give it close to a maximum score. This observation has prompted some to suggest using the inverse of these measures to detect communication signals \cite{rosete2018using}, which correctly gives noise a low score, but only at the cost of giving a very high score to uniform repetitive data \cite{mahon2024towards}, which is not able to convey a complex message either. This is the fundamental problem of detecting the type of complexity we are interested in here: there are several effective methods that define a scale from uniformity at one extreme to random noise on the other, but the data we would like to give the highest score to, such as human language and real-world images, sits in the middle, not close to either extreme. One might try to define a Goldilocks zone where e.g. Shannon entropy is neither too high nor too low, as the location of communicative signals. This is the wrong approach for two reasons. Firstly, there is no principled way to select the boundaries of such a zone. Secondly, it is easy to construct low-complexity, meaningless data that would fall inside it. Consider a bit-string consisting of all 0s, which has minimum Shannon entropy of zero, and imagine replacing the first $p$\% of it with independent uniformly random bits, i.e. coin flips, and varying $p$ from 0 to 100\%. The bit-string remains meaningless throughout, but because its Shannon entropy increases steadily from 0 to the maximum score of 1, it would cross any intermediary zone. 

We claim that, while description length for perfect reconstruction is related to meaningful complexity, it does not tell the full story, and so Kolmogorov complexity and Shannon entropy and related measures are incomplete for quantifying meaningful complexity.

Our solution is to consider descriptions of the data divided into two parts, with only one counting toward complexity. For communication, portion A, the meaningful portion, conforms to a locally compositional structure, meaning a tree-like structure where the leaves are the atomic constituents of the data such as image pixels or text characters. 
Portion A does not (normally) specify the given data exactly, but rather describes an approximation. Portion B, then, fills in the missing details.
Portion B conforms to a default structure that makes minimal assumptions about the data and that we do not regard as meaningful. Both parts have a well-defined length, equal to the number of bits they comprise. We choose the optimal description as that with the overall smallest length (sum of portion A and portion B), and the \ac{lcc} score is then the length of portion A within this optimal description. Effective complexity \cite{gell1996information} employs essentially the same solution, except it generally requires specifying the meaningfully structured features from the outset, and then separately describing these features and the remaining, meaningless, portion. In our framework, on the other hand, we first compute a single optimal description for the entire data, and then analyse it as two separate portions. This means that whether a feature is regarded as being present at all depends on all aspects of the data. The same data could admit either a structured or unstructured description, and which one is selected depends on which one is more efficient. Portion A is analogous to the description of the macrostate of a physical system, and Section \ref{subsec:macro-micro-states} outlines how this framework could be used to provide a more objective characterisation of a macrostate. A comparison can also be drawn to compression: the representation consisting of both portions A and B corresponds to lossless compression, because it can exactly reproduce the given data, while the representation with just portion A corresponds to lossy compression, because it reconstructs an approximation with the less important aspects being ignored. 

The contributions of this paper are:
\begin{enumerate}
    \item a novel approach to measuring complexity by dividing the description into structured and unstructured portions;
    \item a sketch of how this idea can be used to more objectively quantify the entropy and macrostate of a physical system;
    \item the proposal of local compositionality as the structure that, in the case of communicative signals, the structured portion  should conform to;
    \item the development of the \ac{lcc} score, a computable metric for complexity that can be applied to a variety of domains;
    \item the empirical testing of the \ac{lcc} on text, image and audio data, showing that it agrees with our intuitions regarding meaningful complexity;
    \item a demonstration of the potential of the \ac{lcc} score for detecting non-human communication, showing that it can identify the Arecibo message as meaningful, as well as determine its correct aspect ratio.
\end{enumerate}

\section{Local Compositional Complexity (LCC) Score} \label{sec:two-part-descriptions}
The two novel elements of the \ac{lcc} score are the method of dividing descriptions into two portions, A and B, and the choice of local compositionality as the appropriate structure for portion A. The follows sections describe both in detail.

\subsection{Two-Part Descriptions} \label{subsec:two-part-descriptions}
The division of descriptions into parts one and two is similar in spirit to two-part codes in the statistical frameworks of minimum description length (MDL) and minimum message length (MML). Both MDL \cite{rissanen1983universal, grunwald2007minimum} and MML \cite{wallace1968information} descriptions consisting of a model (portion A) and the data input to the model (portion B). The standard division in MDL into model and input is slightly different to the present one. Here, the first part of the description generally contains both something like a statistical model \emph{and} the use of that model to approximate the given data, the second corrects the errors and fills in missing details in that approximation.

For example, if the data consisted of a set of points in $\mathbb{R}^n$, and we considered a fit clustering model as the model of the data, then portion A would consist of the cluster centroids in the fit model and the cluster labels for each data point. This gives an approximation to the given data where every point is replaced with its assigned cluster centroid. Portion B, then, is the residual portion that corrects this approximation to the precise data values. We search for the description that has the overall shortest length, adding together portion A and portion B. Assuming 32-bit precision, each cluster centroid would require 32$n$ bits, and each cluster label would require roughly $\log{K}$ bits, where $K$ is the number of clusters (though this can be further optimized, as discussed below).
For some data points, this two-part description of approximation plus correction, is more expensive than a simple one-part description that specifies data point directly. This simple one-part description can just be describing the floating point values of the data point directly, which would take 32$n$ bits. For each data point, we use whichever is smaller, the two-part or one part description. 
All one-part descriptions fall into portion B, because they do not contribute to any consistent structure or pattern in the data. 
We will see in the experiments in Section \ref{sec:implementation-and-results}, that the total description length for random noise is very high, higher than for data such as speech or real-world images, but almost all the description resides in portion B, so the \ac{lcc} score is close to zero. For uniform, repetitive data, on the other hand, the overall description length is very low, so even when it mostly resides in portion A, the \ac{lcc} score still ends up being low. Note the trade-off of adding more clusters: each additional cluster increases the model length, by 32$n$ in our example, and increases the length of each cluster label, by roughly $\log{k+1} - \log{k} \approx \tfrac{1}{k+1}$, but it means there are more cluster centroids to better cover the data, and so will tend to reduce the average residual cost. The \ac{lcc} score uses the MDL principle and selects the number of clusters that minimises the overall cost. This particular usage of the MDL principle, that of of a criterion for selecting the number of subsets in a partition, has been employed by a number of previous works \cite{kontkanen2005mdl,mahon-lapata-2024-modular, mahon2025parameter}.

\subsection{Entropy, Macrostates and Microstates} \label{subsec:macro-micro-states}
This division of descriptions into a structured portion A and an unstructured portion B is a general framework not specific to communication. Indeed, an analogy can be drawn to describing the state of a physical system in statistical mechanics: portion A corresponds to the macrostate and portion B to the microstate given the macrostate. 

Typically, the choice of what form the macrostate should take is made depending on what `state variables', such as temperature and pressure, the experimenter is interested in or can observe/control. The apparent subjectivity in the choice of what the state variables are, plus the fact that the entropy of the system depends on this choice, has led to the characterization of entropy as an ``anthropomorphic'' property \cite{jaynes1965gibbs}. In our framework, there is a precise criterion for whether to include a particular variable in the macro-description, namely whether it decreases the overall description length of the microstate\footnote{For simplicity, we assume the microstate itself is known, but this could be replaced with the expected description length over the set of possible microstates.}. Once we pick a set of possible state variables, for example all conserved or invariant quantities, we can use this criterion to select, for a particular state, which subset of the variables should be used, and how many bits of precision should be devoted to them. Note, this refers to the accuracy with which we \emph{represent} a quantity, not to the accuracy with which we can measure it. 

For example, suppose temperature was one of the possible state variables, and assume that, for any precisely known temperature, we can calculate a distribution over microstates, then, for a given distribution over temperatures, we can marginalise to calculate another distribution over microstates. Let $q(s | \tau)$ be the distribution over microstates resulting from considering the temperature of state $s$ up to $\tau$ bits of precision. Then, under the optimal encoding scheme as given by the Kraft-McMillan inequality \cite{kraft1949device, mcmillan1956two}, the number of bits needed to select the single correct microstate from this distribution is $-\log_2{q(s | \tau)}$. (See Appendix \ref{app:bayesian-inference-for-tau} for further details.)
Thus, our theory suggests that the correct number of bits of precision to use, $\tau^*$, is given by
\begin{equation} \label{eq:optimal-temperature-precision}
\tau^* = \argmin_{\tau} \tau -\log_2{p_{s}(\tau, s)}\,,
\end{equation}
where $\tau^*=0$ means not including temperature as a state variable at all. This would imply that temperature is not an inherent feature of the system. Just because a feature can be measured and described, does not mean it is correct to do so. For example, the feature `stripe pattern' could be measured and described of certain visual objects even when it is not warranted. One could choose to regard, not just zebras, but all Equidae (including horses) as striped, and say that what appears as a fully white horse, has white stripes on a white background. This would not exactly be false, but it unnecessarily complicates the description of the white horse, and so it seems wrong to include. The two-part framework used in this paper suggests that, if $\tau^*=0$, then temperature is to the system as stripes are to the white horse. In practice, however, other factors, such as measurability, may affect the optimal choice of state variables. In an experiment specifically investigating temperature, there is of course no problem considering it even if $\tau^*=0$.

\begin{figure}
    \centering
    \includegraphics[width=\linewidth]{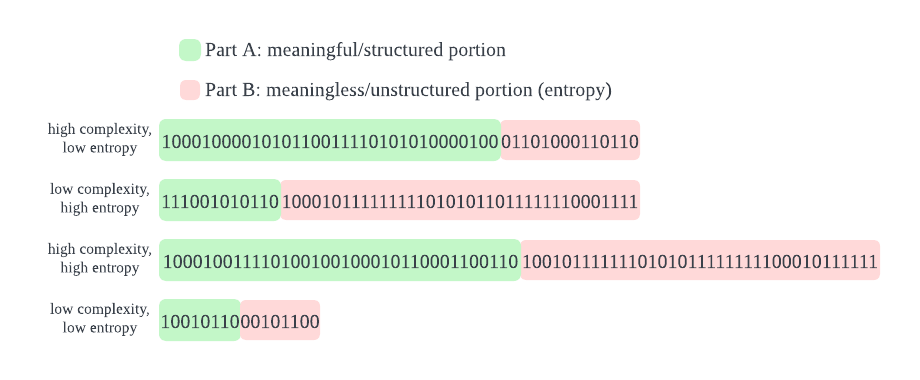}
    \caption{Example, dummy descriptions, of different descriptions of physical systems, showing how high entropy and low complexity can come apart given a varying overall description length.}
    \label{fig:entropy-complexity}
\end{figure}

\paragraph{Entropy vs Complexity}
Entropy and complexity are two important concepts in information theory and science more broadly. 
Given the optimal (i.e. shortest) description of a system, where portion A consists of the macrostate as we have outlined, it is easy to see (shown in the Appendix \ref{app:prop-to-gibbs}) that the expected length of portion B is proportional to the Gibbs entropy. Thus, one could claim that a decrease in Gibbs entropy is equivalent to an increase in complexity, assuming the overall description length remains constant. However, it is likely that this assumption would often be violated in practice: even the same system can have a varying description length as it moves through different configurations. This is shown graphically in Figure \ref{fig:entropy-complexity}. Comparing row two to row one, we see increasing entropy means decreasing complexity, because the two rows happen to have the same total description length. However, rows three and four show that, in the realistic case that total length can vary, high entropy can correspond to either high or low complexity. Entropy is the length of the optimal description of the system that is \emph{not} meaningfully structured, whereas complexity is the length of the portion that \emph{is} meaningfully structured. In general, the two notions are independent.

\subsection{Local Compositionality} \label{subsec:local-compositionality}
Turning specifically to human communication, we claim the appropriate structure for portion A is local compositionality. Compositional means that the structure should be tree-like, in that it represents multiple different parts of the data as `children' of some shared `parent', and that this parent can then be a child of another parent etc. Local means that, assuming the data has a notion of distance between points, so that some parts are close to each other and some far away, then only parts that are close together should share a parent. An example of this structure is grammatical parse trees of natural language text.  We can regard natural language as represented by a locally compositional structure extending from phonemes, through morphemes, syntactic constituents, sentences and discourse elements, where the large majority of compositions (by some theories, all compositions \cite{steedman2001syntactic}) are local. For example, if the words `the' and `cat' are adjacent, then they can be composed into a single grammatical unit `the cat'. Local compositionality is also the structure found in human perception of visual objects, where lines combine to form shapes, and then objects etc \cite{ommer2007learning, lecun2015deep}. 

We do not want a single tree that encompasses all of the data, as multiple different parts of the data should be able to correspond to the same node in the tree: the combination of `the' and `cat' can appear multiple times in the data but should only occur once in the tree. To allow for this, we first describe a tree-like structure, then invoke the nodes it contains to specify the data. We term the former the `model', and the latter, the `index'. 
For human-readable messages, this is the structure that we propose portion A descriptions should conform to. Portion B then consists of an encoding scheme that makes minimal assumptions about the data structure. Examples of these default encodings are detailed below. Generally, they involve encoding each part independently and assuming a normal or uniform distribution. 

In the clustering example from Section \ref{subsec:two-part-descriptions}, the model consisted of the fit cluster centroids, and the index consisted of the sequence of cluster labels. In the physical state example from Section \ref{subsec:macro-micro-states}, the model consisted of a set of macroscopic quantities (state variables), along with the physical equations by which they are related, probabilistically, to microstates, and the index consisted of values for these state variables, up to finite precision. Portion A need not necessarily comprise tree-like structures, in principle one could choose any sort of model. However, for many choices of model, it is possible to find data which gets a very short description, but appears random and meaningless to humans. An example is given in Appendix \ref{app:bad-model}. In order to align with what is meaningful to humans, we propose that tree-like structures, specifically local compositionality, is an appropriate choice of model type. 

The models used to compute the \ac{lcc} score are not assumed to be like the process that, in some sense, `really' generated the data. The framework is just that we are given a piece of data, with no prior ideas about it, and then must search among a set of possible representations for the most efficient one. 
This is in keeping with the MDL philosophy of focussing on finding efficient descriptions, rather than a true distribution or generating function \cite{grunwald2007minimum}. The \ac{lcc} score is therefore not a property of a distribution but of a given piece of data or dataset. It generally increases as the amount of data is larger, which reflects the fact that it is a version of information content and that more data can generally convey more information.

We do not have a formal proof that the \ac{lcc} score will always give a high score to data if and only if it contains a human-readable message. A proof is given in \cite{mahon2024minimum} that, for their very similar method, white noise images will, with high probability, get a score of zero, but even that is not a formal guarantee against false positives, because white noise does not constitute all non-meaningful-to-human images. It is difficult to see how any such metric could have a formal proof of correctness because we do not have a formal account of what makes something a human-readable message--in fact one way to interpret the LCC score is as an attempt at such a formalisation. Instead, we show empirically, in the following sections, that it gives a high score to a set of things we know to be meaningful to us, and a low score to a set of things we know not to be.


\section{Implementation and Results} \label{sec:implementation-and-results}
We now describe more concretely the implementation of the \ac{lcc} score for data from different domains. We then present experimental results showing that it consistently gives a low score to random data and to repetitive/uniform data, and that the only data that gets a high \ac{lcc} score is that which we know to be meaningful in the sense that it carries a human-readable message. 

\subsection{Discrete Data} \label{sec:discrete-data}
In the domain of strings, such as natural language text, the model is a codebook mapping single characters to sequences of characters, and the index is a string 
composed of characters that appear as keys in the codebook, along with a special character $x$, indicating `unspecified'. The data is specified by proceeding from the top of the codebook, and replacing all occurrences of each key that appear in the index string with the corresponding value, then replacing all $x$ characters with specific values as given in a third string, the residual string. The reverse process of replacing all occurrences of a subsequence with an index to the code, we refer to as `aliasing'. This conforms to the local compositionality structure, and allows an efficient approximate search for the optimal representation (see Appendix \ref{app:algs}). The fact that values in the dictionary can be multiple characters long means the tree is not restricted to be binary and can express $n$-ary relations, as discussed for example in \cite{johnson-13}. The complexity score is the total number of bits needed to represent the codebook and the index string, under the optimal encoding, ignoring the residual string. 

Adding more entries to the codebook reduces the number of indices in the input string, causing a reduction in cost, but increases the average length of each index and size of the codebook itself. For random sequences of characters, the decrease will almost always be outweighed by the increase, meaning almost nothing is added to the codebook and all the information is contained in the residual string, so the \ac{lcc} score will be low. Repetitive uniform data, on the other hand, can be represented accurately by only a few codebook entries, so the codebook will be small and will admit efficient indexing, so it will also get a low score. 

\subsubsection{Discrete Worked Example}
Let $\mathcal{A} = \{a,b,c\}$, and 
\begin{gather*}
    S = ccabacbbbcaacbcccbcaaacbbcccabbbaacbaaabcabbbcabbbcacbb\,.
\end{gather*}
Then, one possible encoding is given by 
\begin{gather*}
    C = ((g, ace), (f, cb), (e, bb), (d, aa)) \\
    I =  xxxxgxxdfxxfxdgxxxxexdfdxxxxexxxexxg \\
    X = ccabbcccccccababcabcabc\,.
\end{gather*}
If we begin with $I$, then replace each codebook character with the corresponding string, followed by replacing the $x$'s with the residual string $X$, we get the following sequence of strings:
\begin{align*}
    I = &xxxxgxxdfxxfxdgxxxxexdfdxxxxexxxexxg \\
    &xxxxacexxdfxxfxdacexxxxexdfdxxxxexxxexxace \\
    &xxxxacexxdcbxxcbxdacexxxxexdcbdxxxxexxxexxace \\
    &xxxxacbbxxdcbxxcbxdacbbxxxxbbxdcbdxxxxbbxxxbbxxacbb \\
    &xxxxacbbxxaacbxxcbxacbbxxxxbbxaacbaaxxxxbbxxxbbxxacbb \\
    &ccabacbbbcaacbcccbcaaacbbcccabbbaacbaaabcabbbcabbbcacbb = S\,.
\end{align*}
The number of bits needed to represent $S$ under this encoding is then $L(C) + L(I) + L(X) = 102.84 + 59.81 + 33.90 = 196.55$ and the computed \ac{lcc} score is then $L(C) + L(I) = 102.84 + 59.81 = 162.65$.\footnote{This particular encoding is the one found by our search algorithm, slightly simplified for demonstration purposes. However, this search is approximate only, so we cannot guarantee that the above encoding is optimal for the present example.}


\subsubsection{Complexity of Natural Language Text}
Table \ref{tab:text-results} shows the score for the first 7500 characters of 10 random Wikipedia articles (precise list in Appendix \ref{app:wikis-list}) each of different natural languages: English, German and Irish, along with the score for other types of strings. The total height of the bar for each string type is its total description length. This is broken down as model cost (dark green), index cost (light green) and residual cost (orange), with the \ac{lcc} score (written in green text) being the sum of the first two. Random strings, `rand', get a score of 0, despite requiring the highest number of bits to represent exactly, because the cost resides entirely in the residual portion. Simple, artificial English, `simp-en', consisting of the same five English words selected randomly, gets a significantly lower score, and very repetitive strings, `repeat[2510]', consisting solely of repetitions of the same random string of length 2, 5 and 10 respectively, get a very low score. The slightly lower score for Irish compared to German and English is expected, given that its alphabet uses only 18 letters, vs 26 for English and 30 for German. One may then expect German to receive a proportionally higher score than English. The fact that it instead receives only a slightly higher score than English is, we expect, due to some feature of the morphology, or perhaps syntax, of the three languages. One possibility is consonant clustering, which occurs more strongly in German and Irish than in English, so e.g. “sch” in German or “bhf” in Irish contains less information than the sum of the constituent letters, thus lowering the scores for these languages.

\begin{figure}
    \centering
    \includegraphics[width=\linewidth]{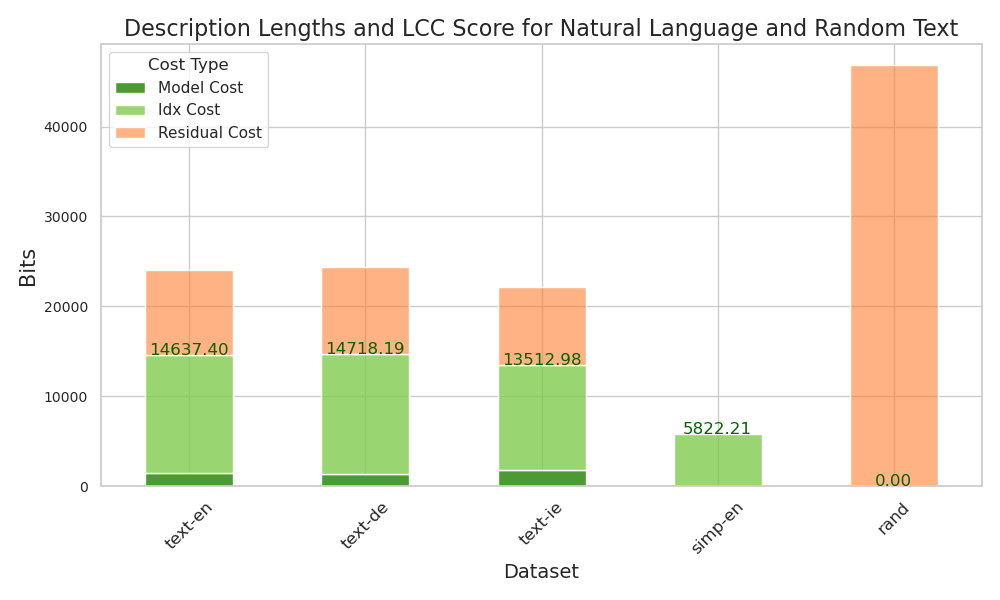}
    \includegraphics[width=\linewidth]{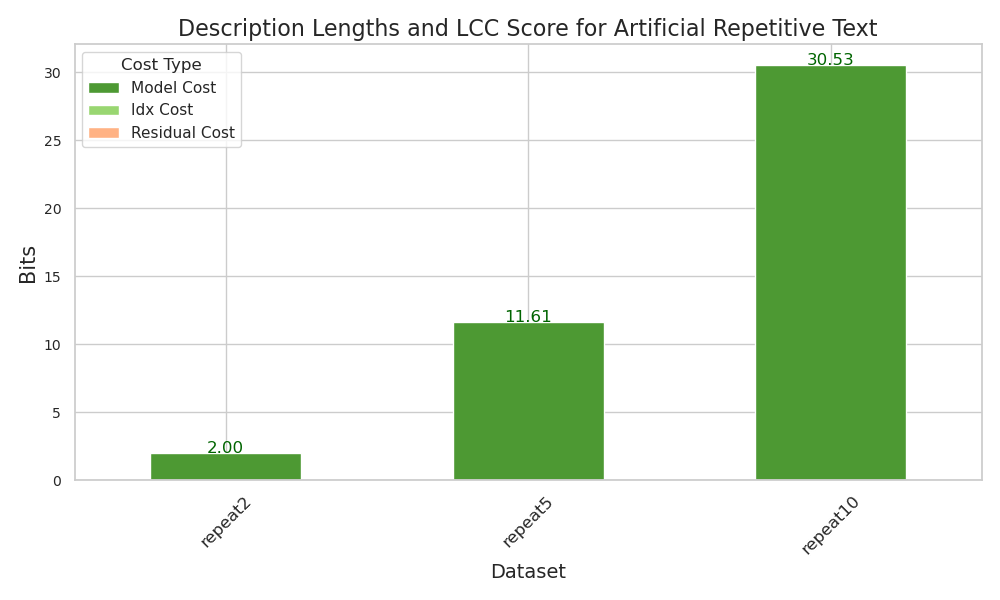}
    \caption{The description cost, in bits, for text in different natural languages, along with simplified English sentences and random text (top) and for artificially generated repetitive text (bottom). This is broken down into the cost to describe the model, the cost to index into that model to describe part of the data (`idx cost'), and the residual portion not accounted for by the model. Portion A of the description comprises the model cost and the index cost. This constitutes the \ac{lcc} score and is marked in green. Random text has the highest total cost but the lowest meaningful cost.}
    \label{fig:nl-text-results}
\end{figure}

\FloatBarrier

\subsection{Continuous Data} \label{subsec:continuous-data}
Unlike discrete data, such as text, continuous data, such as images and audio, allows two different values to be arbitrarily close to each other. 
For continuous data we cluster progressively larger localised regions, e.g. image patches, using a multivariate \ac{gmm}. The representation model then consists of a sequence of \acp{gmm}, specified by the means and covariances of each cluster. The index consists of an $n$-dimensional 
tensor (for images $n=2$, and for audio $n=1$), where each location is either an index to a cluster of the \ac{gmm} or the special symbol $x$. 
The residual consists of the encoding of each point in the input, with respect to its assigned cluster in the \ac{gmm},
plus the 
values of each $x$.

A similar idea was used by \cite{mahon2024minimum}, except that their method only applied to the image domain, and included a complicated entropy calculation that was not justified in a theoretical framework and that did not transfer to other data types. 

Further details, including a full formal description, for the continuous calculation of \ac{lcc} score is given in Appendix \ref{app:continuous-full-description}.

\subsubsection{Continuous Worked Example}

\FloatBarrier
\begin{figure}
    \centering
    \includegraphics[width=0.5\linewidth]{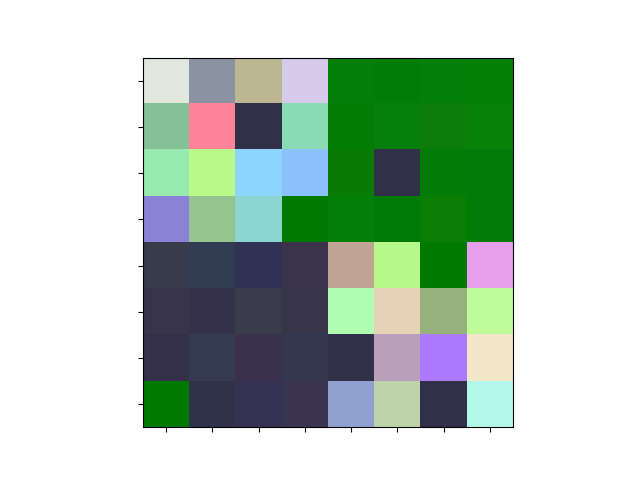}
    \includegraphics[width=0.5\linewidth]{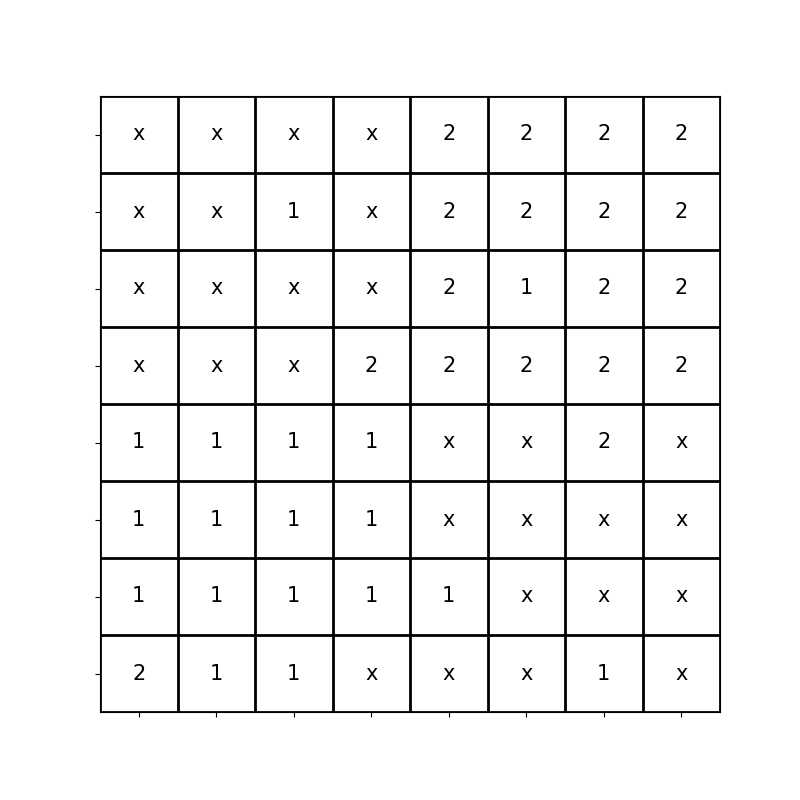}
    \includegraphics[width=0.5\linewidth]{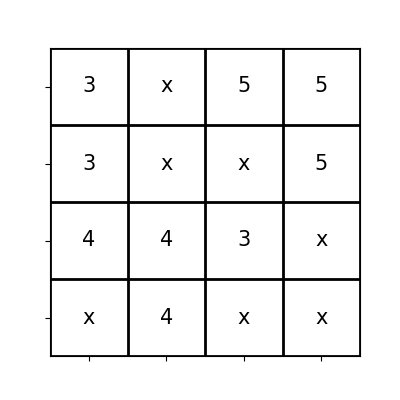}
    \caption{Top: Toy example image. Middle: The MDL cluster labels for this image on the first level. One this level, two clusters are found, corresponding to blue-ish and green-ish pixels, and the meaningful portions, i.e. those pixels assigned a cluster centroid, and not marked with 'x', are mostly in the bottom left and top right. Bottom: The MDL cluster labels on the second level. One this level, there are again two clusters, now corresponding to green-ish and blue-ish patches, and the meaningful portions are entirely in the bottom left and top right.}
    \label{fig:img-example}
\end{figure}

\begin{figure}
    \centering
    \includegraphics[width=\linewidth]{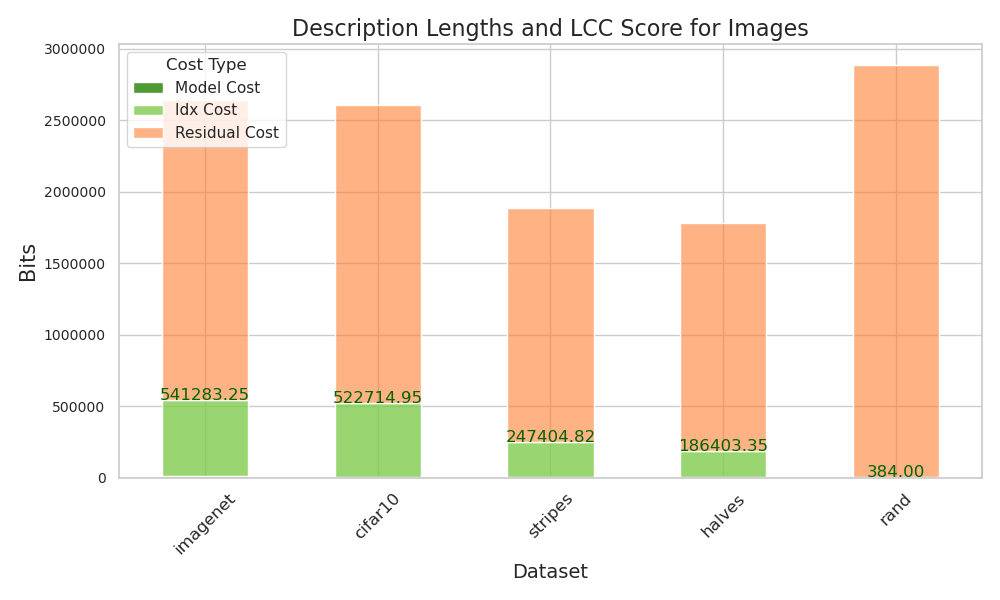} 

    \hspace{4em}
    \includegraphics[width=0.2\textwidth]{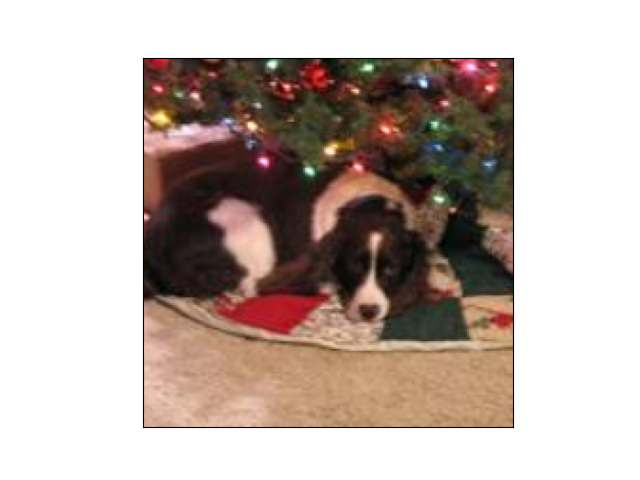} \hspace{-2em}
    \includegraphics[width=0.2\textwidth]{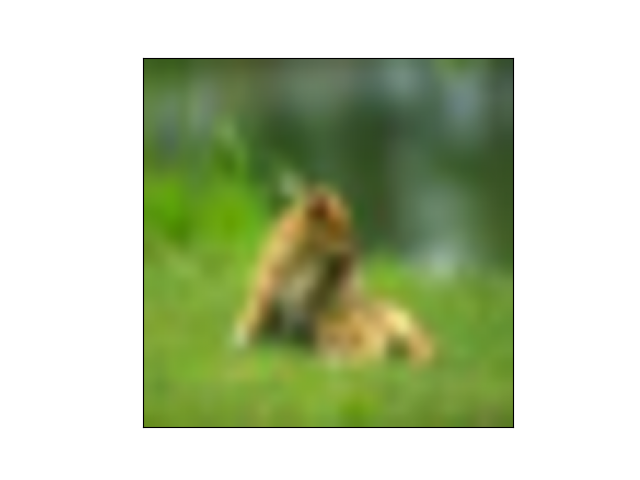} \hspace{-2em}
    \includegraphics[width=0.2\textwidth]{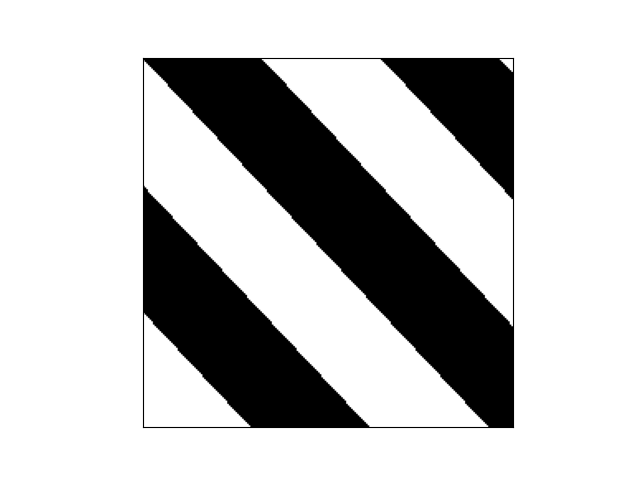} \hspace{-2em}
    \includegraphics[width=0.2\textwidth]{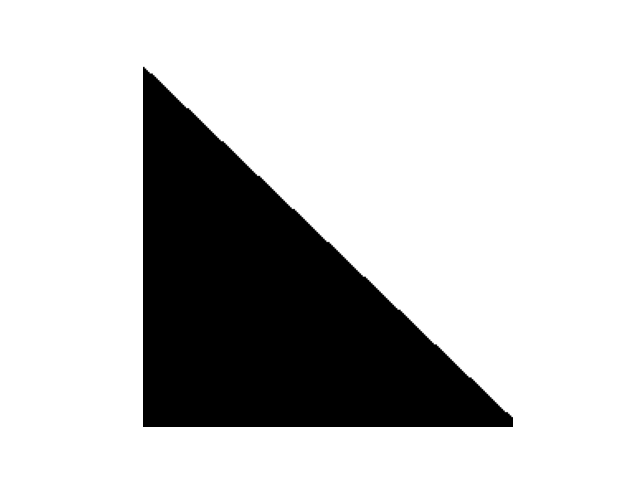} \hspace{-2em}
    \includegraphics[width=0.2\textwidth]{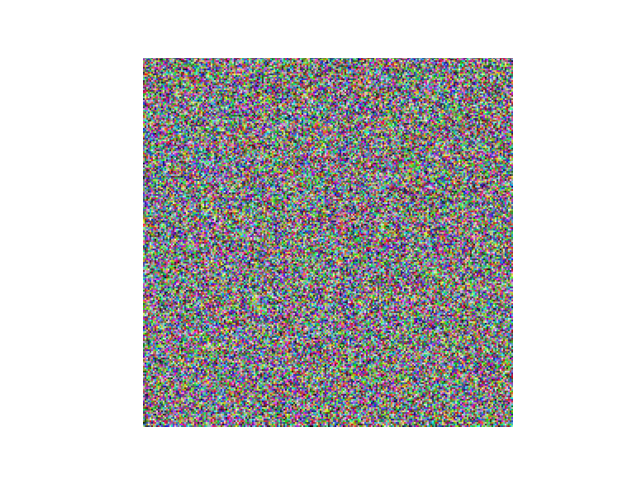}
    \caption{The mean description cost, in bits, over 100 randomly sampled images of various types (one example displayed below each column). The cost is broken down into that for the model (`model cost'), for indexing into that model to describe part of the data (`idx cost'), and for the residual portion not accounted for by the model (`residual cost'). Portion A of the description comprises the model cost and the index cost. This constitutes the \ac{lcc} score and is marked in green. Random images have the highest total cost but the lowest meaningful cost. For readability, the displayed heights for the residual portion are reduced by a factor of 5.}
    \label{fig:image-results}
\end{figure}

Here we present a simplified worked example for the continuous domain. Suppose the input image is as in Figure \ref{fig:img-example} (top). The first step is to run MDL clustering on the set of pixel values across the whole image. In this case, the result is two clusters corresponding to blue-ish and green-ish pixels respectively. Some pixels are not assigned any cluster, and are instead encoded directly as part of the residual, because they do not fit efficiently into a pattern with the rest of the data. Writing $x$, for such pixels, we get the assignment as shown in Figure \ref{fig:img-example} (middle). Most of the pixels in the blue cluster are in the bottom left and most of those in the green cluster are in the top right. In total, there are 19 blues, 18 greens and 27 zeros. The Shannon information content of this distribution of labels is 
\[
-19\log_2\frac{19}{64} - 18\log_2\frac{18}{64} - 27\log\frac{27}{64} \approx 33.29 + 32.94 + 33.61 = 99.59\,.
\]
The model cost, meanwhile, comprises the bits for the RGB values for each of the two clusters. At 32-bit precision, this gives $2 \times 3 \times 32 = 192$. The next step is to cluster patches in the image. By `patch', we here mean a $2\times 2$ region of pixels, see Appendix \ref{app:continuous-full-description} for a full description of the computation of patch cluster labels. In the full method, we consider overlapping patches, but for simplicity here, we will show non-overlapping of size $2 \times 2$. This would give three clusters, those mostly composed of 0s, 1s, 2s. The result is as in Figure \ref{fig:img-example} (bottom).\footnote{These are for illustrative purposes only and may differ from those output by our method.} There are seven 0s, three 3s, three 4s and three 5s, giving information content
\[
-3\log_2\frac{3}{16} - 3\log_2\frac{3}{16} - 3\log\frac{3}{16} - 7\log\frac{7}{16} \approx 7.25 + 7.25 + 7.25 + 8.35 = 30.08\,.
\]
Here, there are three cluster centroids, which takes $3 \times 3 \times 32 = 288$. The total score for this image is then 192 + 99.59 + 288 + 30.08 = 609.67. Because this toy image is small, the cost is dominated by the model cost, but for larger images, it is dominated by the index cost, as can be seen in Figure~\ref{fig:image-results}.

\FloatBarrier
Table \ref{fig:image-results} shows the \ac{lcc} score for various types of images: Imagenet and Cifar10 are two datasets of natural images depicting a single, identifiable object. `halves', and `stripes' are two simple uniform types of image we create, the former are half black and half white, with the dividing line being at various angles, and the latter consists of black and white stripes of varying thickness and angles. We also include white noise images, `rand'. The natural images consistently get the highest score, and a very similar one across the two datasets, Imagenet and Cifar10; the simple uniform patterns get a moderate score, and random noise gets the lowest score of close to zero, even though the total cost is the highest of all image types. 
It is also striking that the fraction of the cost occupied by the residual description is much higher for images than for text. In order to make visible the differences in the \ac{lcc} score portion, the residual portion of the bars in Figure \ref{fig:image-results} has been reduced by a factor of 5. The actual ratio of residual cost to \ac{lcc} score (model cost plus index cost), is 20:1, even for the real-world images in Imagenet and Cifar10, and even higher for the other images. For natural language text, in contrast, this ratio is over 1:3. This is consistent with the conception of text as information dense: if we regard `information density' as the ratio of the length of the meaningful portion of the description to the total length of the description, then our results find natural language text to be $\sim$60 times as information dense as real-world images. 

\begin{figure}
    \centering
    \includegraphics[width=\linewidth]{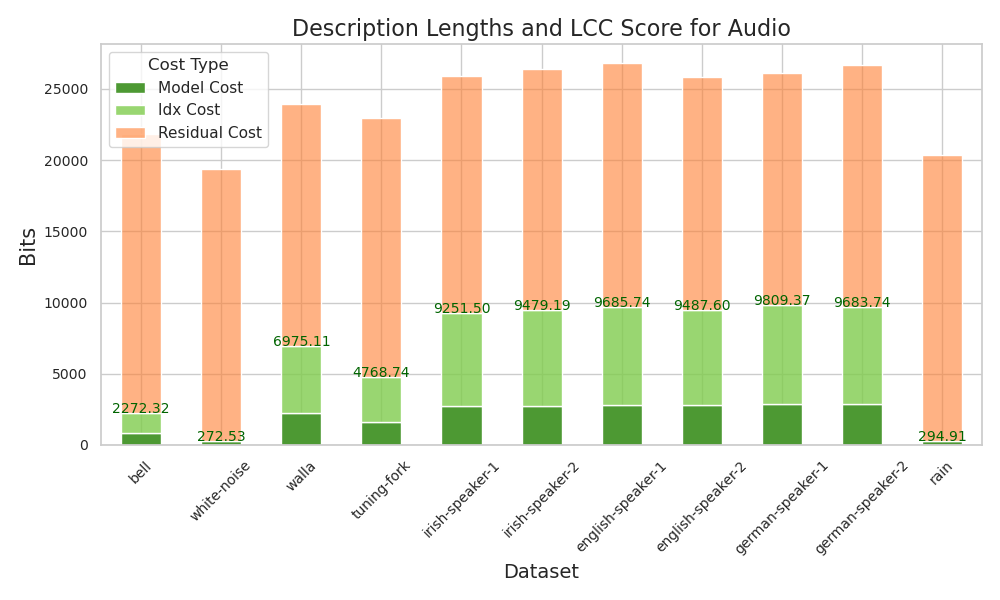}
    \caption{The description cost, in bits, for different types of audio signals, broken down into the cost to describe the model, the cost to index into that model to describe part of the data (`idx cost'), and the residual portion not accounted for by the model. For readability, the displayed heights for the residual portion are reduced by a factor of 5. Portion A of the description comprises the model cost and the index cost. This constitutes the \ac{lcc} score and is marked in green.}
    \label{fig:audio-results}
\end{figure}

Our method can be applied to audio in the same way as to images, except that we first convert the audio signal to a spectrogram. 
Table \ref{fig:audio-results} shows the resulting scores for various types of audio: human speech in English, Irish and German; random noise, sampled from a uniform distribution 
; simple, repetitive signals, a tuning fork and a continuously ringing electronic bell; and some examples of ambient noise: rainfall and muffled crowded human speech (walla). We see the same patterns as before: human speech gets the highest score, random noises have a high total description length but almost all of this is in portion B, so they get a very low score, and simple repetitive signals have a score in between the two. It is striking that the three natural languages get such a similar score to each other, which is consistent with the general principle that, phonetically, all human languages are roughly equally efficient at conveying meaning \cite{liversedge2016universality, hawkins2014cross, rubio2021speakers}.

\subsubsection{Use in Compression}
\begin{figure}
    \centering
    \includegraphics[width=0.45\linewidth]{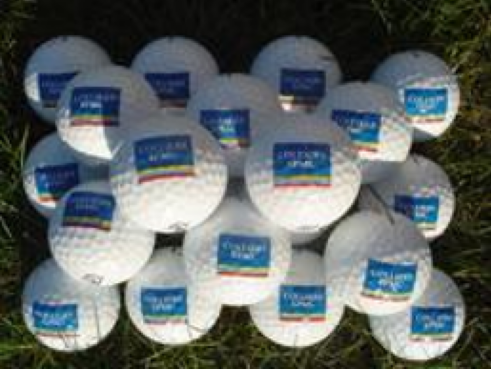}
    \includegraphics[width=0.45\linewidth]{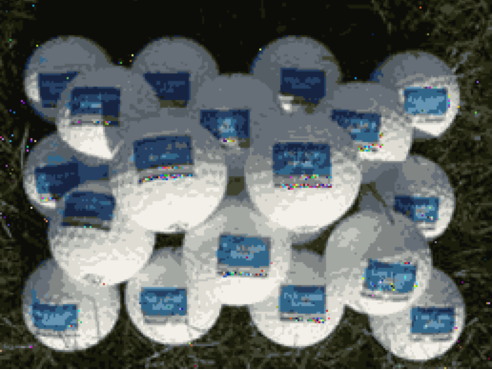}
    \includegraphics[width=0.45\linewidth]{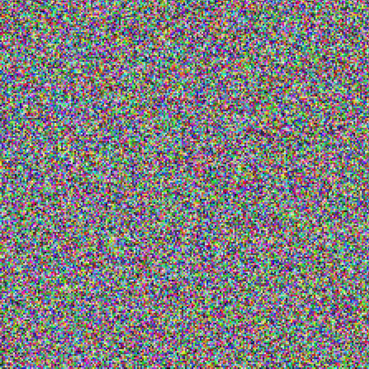}
    \includegraphics[width=0.45\linewidth]{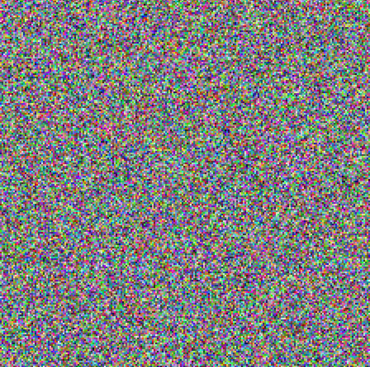}
    \caption{Example of a original images (left) and the reconstructions resulting from using only portion A of the \ac{lcc} representation as a form of lossy compression (right). The top row shows a complex real-world image, and has a compression ratio of 25, c.f. JPEG's $>200$. Note the random colours on the portions that we deemed outliers. The bottom row shows a random noise image, and has a compression ratio of 114, c.f. JPEG's 32.}
    \label{fig:compression-examples}
\end{figure}

The problem of efficiently representing data is studied in the context of compression, which is the process of encoding data to reduce its size by exploiting redundancy and patterns. 
The entire description, portions A and B, constitutes a sort of lossless compression, while discarding portion B would constitute a lossy compression. The original image can be approximately reconstructed from portion A alone, by replacing the information in portion B with random samples from the corresponding cluster distributions. Figure \ref{fig:compression-examples} shows such reconstructions for an example real image and noise image. 

The \ac{lcc} representation has not been optimized for compression. Specific compression algorithms, such as JPEG, carefully preserve information most salient to the human eye. For the real-world image, JPEG provides a much better quality-compression trade-off than the reconstruction in Figure \ref{fig:compression-examples}, reaching over 200 before a similar degradation in quality. For the noise image, the \ac{lcc} compression ratio is 114, while the JPEG is only around 30 for a similar quality. Carefully examining the noise reconstruction from the \ac{lcc} method, the pixel values are actually completely different, yet the \ac{lcc} representation correctly identified that these differences are not structurally meaningful, giving a large compression ratio. The intention of the \ac{lcc} representation is not compression, but rather to be theoretically optimal among a certain class of representations, enabling reasoning and quantification of meaningful structure. Clearly, JPEG affords better compression overall than the simple compression method shown in Figure \ref{fig:compression-examples}. However, it is interesting that it provides a reasonable compression as a by-product, and this may suggest a direction for future work, especially in the context of compressing noisy data.

\subsection{Arecibo Message} \label{subsec:arecibo-results}
If the earth were to receive a communication signal from elsewhere in the universe, even before decoding what it meant, an interesting and important question is whether we would be able to detect that it contained a message at all. We cannot assume the language of this message would be especially similar to any found on Earth. However, if we make no assumptions about its possible structure, then any bit-string could be associated with any meaning and the problem would be impossible. We need to assume enough about a possible alien language to have a chance of detecting it, but not so much that we rule out those that would be very unfamiliar to us but still potentially decipherable. The assumptions of locality and compositionality are reasonable choices in this respect, and the results presented so far already show that the present method has a strong ability to detect signals of this sort. 

\begin{figure}
    \centering
    \includegraphics[width=0.3\linewidth]{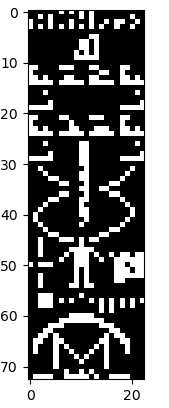}
    \includegraphics[width=0.3\linewidth]{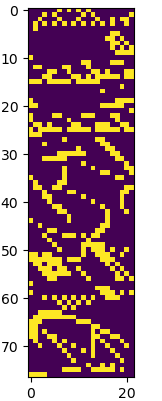}
    \includegraphics[width=0.6\linewidth]{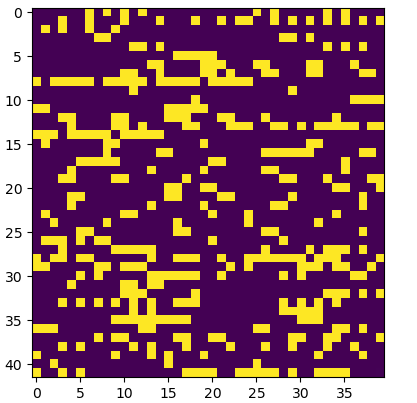}
    \includegraphics[width=0.6\linewidth]{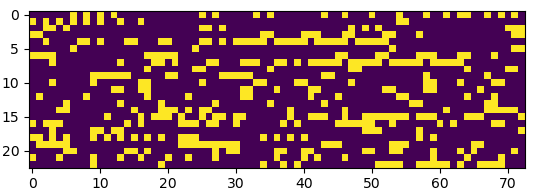}
    \caption{The bitmap image of the Arecibo message, shown at four different aspect ratios: the intended shown in black and white at $73 \times 23$ (top left), one very close to the intended at $77 \times 22$ (top right), the most square-like, at $42 \times 40$ (middle), and the inverse of the intended, at $23 \times 73$ (bottom). The \ac{lcc} score is high for the top two (highest for the top left) and low for the bottom two.}
    \label{fig:arecibo-img}
\end{figure}

\begin{figure}
    \centering
    \includegraphics[width=0.8\textwidth]{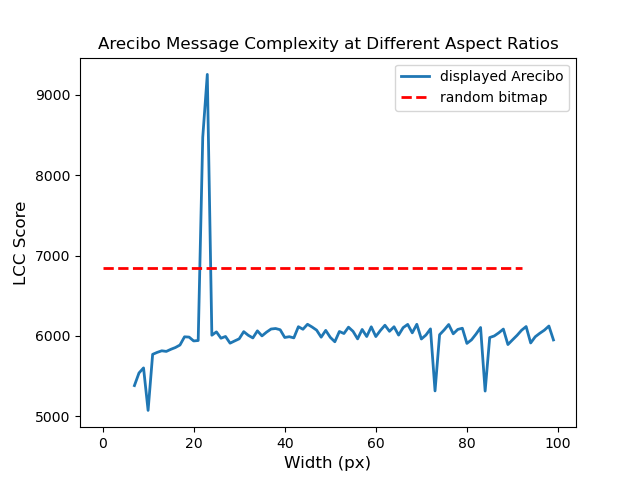}
    \caption{The \ac{lcc} score for the Arecibo message at different aspect ratios (blue line), relative to the maximum score that could be given to a random bit-map of the same length (red dotted line). The correct aspect ratio of 23 $\times$ 73, along with a similar one of 22 $\times$ 77, score much higher than others and than the random benchmark.}
    \label{fig:arecibo-plot}
\end{figure}

Now, we directly consider the case of interplanetary communication, by computing the \ac{lcc} score for the Arecibo message, a sequence of 1679 bits broadcast by a group of Cornell and Arecibo scientists, including Carl Sagan and Frank Drake, in 1975 \cite{arecibo75}. When depicted as a $73 \times 23$ bitmap image, as shown in Figure \ref{fig:arecibo-img} (top left), it shows a series of simple shapes and patterns designed to convey information about Earth and Humankind. When depicted with aspect ratio that differs by more than +/-1 from this, these shapes are lost and it becomes unintelligible to human eyes. Figure \ref{fig:arecibo-plot} shows the \ac{lcc} score given to the image for varying aspect ratios.\footnote{The signal length of 1679 was chosen to be semi-prime so it only exactly factors in two unique ways as $73 \times 23$ or $23 \times 73$. We could imagine that the receiver is unsure of the exact start and end point, which introduces error in the possible length and greatly increases the set of possible aspect ratios. In any case, the problem of singling out the correct one of $73 \times 23$ based only on the resulting image contents is an instructive one for the present purposes. For different aspect ratios, we add the minimum number of bits to the end to allow the number to factor in that ratio, with values independent and random with the same on probability as the rest of the signal. This is a small number of additional bits, often only 1 is needed; the average is 22, which is $\sim 1.3\%$ of the total. }

The red dotted line shows the maximum score that could be given to a random bit-string (obtained at Bernoulli p=0.5). The correct aspect ratio gets the highest score, of over 9000, and clearly exceeds the threshold, while almost all others do not. Note that we are representing the message as a continuous image, meaning the total number of bits is $32 \times 1679 = 53728$. The second highest, and also above the red line, is 1 pixel in width away from the correct one, the ratio of $77 \times 22$ (with 15 random bits at the end). The image at $77 \times 22$, shown in Figure \ref{fig:arecibo-img} (top right), is in fact still somewhat intelligible. The human shape midway down the image, and the antenna shape at the bottom, have merely been slanted to the left. Given that the \ac{lcc} score does not know what a human or antenna or any other particular shape is, 
it is reasonable that it considers the $77 \times 22$ aspect ratio to be similarly meaningful to $73 \times 23$. Indeed, it is striking that, without any of this information on the shape of familiar objects, or previous examples of such objects, the \ac{lcc} score can identify that the signal for the Arecibo message is likely to contain a human-readable message, and the particular way the data should be represented for this message to be human-readable. If humans received an extra-terrestrial message in the same form as the Arecibo message, we could not expect it to depict objects we recognize, such as human bodies and radio antennas, but if it was organized in a locally compositional structure, then the \ac{lcc} score may be able to identify it meaningful, as well as the aspect ratio in which to read it.

\cite{zenil2023optimal} also proposed a complexity measure, applied to the Arecibo message image at different aspect ratios. Their measure gave the \emph{lowest} score to the correct aspect ratio, in contrast to ours which gives the highest. This is because
the overall description is indeed shortest for the correct aspect ratio. However, in the \ac{lcc} framework, a higher fraction of that description is structured, and so the overall score is higher. This again highlights how the \ac{lcc} score is measuring meaningful structure, rather than, as existing methods do, approximating Kolmogorov complexity. As I have argued, approximating Kolmogorov complexity is not able to make the correct three-way distinction between simple, meaningful and random. For example, a method like that of \cite{zenil2023optimal} will give the lowest score to uniform data.

\section{Discussion} \label{sec:discussion}
The first main idea behind the \ac{lcc} score is to divide an information-theoretic description into two parts, and to take the length of only the former, structured part as a measure of complexity. This is similar to some purely theoretical computer science work that seeks to modify Kolmogorov complexity to not give a high score to random strings. `Sophistication' \cite{koppel1987complexity,vitanyi2006meaningful} and `effective complexity' \cite{ay2010effective,gell1996information}, propose to divide the generative algorithm for a given string into a section that describes a set of which it is a typical member, and a portion that selects it from within that set, and take the length of the former as the complexity. Our method differs in that it determines the boundary between the structured and unstructured section dynamically, as part of a single optimisation of the overall description length. Additionally, while these existing methods are purely theoretical and often uncomputable, ours can be implemented to actually compute accurate complexity scores for a variety of data, as shown by our results.

The second aspect of the \ac{lcc} score, local compositionality, has a connection to \ac{at} for measuring molecular complexity \cite{marshall2021identifying,cronin2006imitation}. 
\ac{at} measures the complexity of a molecule as the number of steps needed to assemble it from atomic constituents, allowing reuse of previously assembled intermediary structures
at the cost of only a single step. This is analogous, in our framework, to aliasing, where we pay the cost of a substring as it appears in the codebook, and then, every time it appears in the input, pay only the cost of indexing that entry. \ac{at} says you can reuse any substring as a single step, which is equivalent to assuming constant lookup time in the codebook, however strictly speaking,
the amortized lookup space (and hence, time) is always bounded below by the Shannon entropy of the codebook entries in the input string \cite{shannon1948mathematical}, so can grow arbitrarily. In our framework, the index length is proportional to the substring’s negative log frequency in the input string, which means we respect this lower bound. 

Similarly, when AT is applied to collections of molecules, some of which may have been assembled by chance, it assumes that any appearing more than once could not have been due to chance \cite{sharma2023assembly}. Specifically, the $-1$ in the numerator of Equation 1 in \cite{sharma2023assembly} is so that molecules contribute to the assembly index of the ensemble if and only if they have two or more copies. This may work to exclude chance occurrences in the case of molecular complexity, where the space of possibilities is far larger than the given set, but in general
items can occur many times due to chance alone. Rather than a threshold of 1, our framework produces a threshold strictly greater than 1, given by 
\begin{equation} \label{eq:str-rand-threshold}
    1 + \frac{\sum_i \log_2 \frac{f_i}{f_i'}}{l}\,,
\end{equation}
where $l$ is the description length of the substring, and $f_i$ and $f_i'$,  are, respectively, the the frequency of the $i$th character in the non-aliased portion before and after aliasing (derivation in Appendix \ref{app:rand-thresh-deriv}). In the case of random strings, as measured in Table \ref{tab:text-results}, we observe that substrings of length up to $\sim10$ characters often occur multiple times by chance alone. If we used a fixed threshold of 1, these substrings would mistakenly be regarded as meaningful. However, we also observe that the number of times they occur is always less than the threshold in \eqref{eq:str-rand-threshold}, so our method still correctly identifies that they did not arise by any meaningful process. The inverse dependence on $l$ means that, for larger substrings, \eqref{eq:str-rand-threshold} approximately recovers the threshold of 1, as used by \cite{sharma2023assembly}. 

Entropic slope \cite{MCCOWAN1999409}, is another method aiming to recognise communicative data. It measures the decrease in entropy of each term in a sequence of symbols as you condition on more and more previous terms. That is, in the terminology of language modelling, the dependence of the perplexity on the size of the left context (though entropy is simply calculated empirically, without any specific language model). A score close to -1 indicates high complexity, because it means the symbols individually are non-uniform, but yet can be predicted more and more accurately with more and more left context. Random strings see no drop in entropy with increasing left context, so have a slope of zero, while repetitive sequences have a low entropy for even a small left context, so cannot drop much further. Thus, in the text domain, entropic slope is able to make the three-way distinction between random, uniform and meaningful. However, it has also been applied to the Arecibo message, and here it gives it a relatively low score \cite{doyle2011information}. The unique feature of the \ac{lcc} score is that it is able to identify meaningful data across a variety of domains, as well as its clear theoretical foundations in minimum description length and Occam`s razor. To our knowledge, it is the first method that has been demonstrated capable of detecting meaningful complexity across various domains including in the Arecibo message.

\section{Conclusion}
This paper described a way to measure the amount of meaningful complexity, of the sort found in human communication, present in a piece of data. First, we described a general framework for constructing two-part descriptions that can be used to quantify any sort of complexity or structure in data. This framework connects to the concept of complexity in statistical mechanics, implying that complexity should not be equated with low entropy, and suggesting a method for a more objective characterization of the macrostate of a physical system. Then, we proposed the particular structure of local compositionality as appropriate for modelling human communication and perception. Coupled with our framework for two-part descriptions, this led to a metric for meaningful complexity, termed the \ac{lcc} score. We showed experimentally that the \ac{lcc} score correctly distinguishes meaningful data from both simple uniform data and random noise, across the domains of text, images and audio. Finally, we applied it to a message broadcast from Earth constructed by human cosmologists, and showed that it identified that this signal contained a message, as well as the appropriate form in which to read the message. This suggests a use for the \ac{lcc} score in the study of non-human communication.

\bibliography{bibliography}


\appendix

\section{Proportionality to Gibbs entropy} \label{app:prop-to-gibbs}
Under the Shannon-Fano optimal encoding scheme, the description length of the $i$th microstate, which has probability given by $p_i$, is $-\log_2{p_i}$. Then the expected code length across all microstates is 
\begin{align*}
    &\sum_i p_i (-\log_2{p_i}) = \\
    = & \sum_i p_i (-\log_2{e} \ln{p_i}) = \\
    = & -\log_2{e} \sum_i p_i \ln{p_i} \propto \\
    \propto & k_B \sum_i p_i (-\ln{p_i}) \,.
\end{align*}
Indeed, a common description of the Shannon-Fano optimal encoding is that the expected code length equals the informational entropy, and this is proportional to the Gibbs entropy by a change of base from 2 and $e$, and the addition of the Gibbs constant $k_B$.

\section{Expression Distribution Overs Microstates Given Temperature} \label{app:bayesian-inference-for-tau}
Let $p(s, T)$ be the joint distribution of microstates and temperatures for the given system. In general, if we have a distribution over temperatures, we can marginalise to get a distribution over states:
\[
p(s) = \int_{t \in \mathbb{R}^+} p(s|T=t)\,.
\]
For any precision $\tau$, denote by $I(t,\tau) \subset \mathbb{R}$ the interval of possible values that have temperature $t$ when expressed with $\tau$ bits of precision. Let $m$ be the measured temperature of the system. Then, assuming a uniform distribution over all possible values of temperature corresponding to t given precision, we can define
\begin{gather*}
    q(s|\tau) = \int_{t \in I(m, \tau)} p(s|T=t)\,.
\end{gather*}
Due to the behaviour of floating point numbers, the size of the interval $I(t,\tau)$ depends on the temperature $t$. For larger temperatures, more bits will be devoted to the exponent and fewer to the mantissa, so the interval will be larger. If, to simplify, we assume a max possible temperature of 1, then the interval will be of length $2^{-\tau}$, and we have
\begin{gather*}
    q(s|\tau) = 2^{\tau} \int_{t \in [m - 2^{-(\tau+1)} , m + - 2^{-(\tau+1)})} p(s|T=t)\,.
\end{gather*}

\section{Full Formal Description of \ac{lcc} score}
\subsection{Formal Description for Discrete Data} \label{app:full-description-discrete}
Let $\mathcal{A}$, the alphabet, be a set of characters, and let $\mathcal{A}^*$ denote the set of all strings of symbols from $\mathcal{A}$. Let $u: \mathcal{A}* \rightarrow P(\mathcal{A})$ be the function that returns the set of all unique characters in a given string. Define a cost function
\begin{gather*}
L: \mathcal{A}^* \rightarrow \mathbb{R}+ \\
L(S) = \sum_{c \in u(S)} - \log_2{\frac{\sum_{c' \in S} \mathbbm{1}(c'=c)}{|S|}}\,.
\end{gather*}
This cost returned by $L$ is the compressed size of $S$ under a theoretically optimal compression method that decodes symbols individually, such as Huffman coding, but ignoring integer rounding. 

Define character-replacement as a function 
\[
R: \mathcal{A} \times \mathcal{A}^* \rightarrow \mathcal{A}^*\,,
\]
where $R(c, s, S)$ is the string that results from replacing all occurrences of character $c$ in $S$ with string $s$. Similarly, define sequential-replacement as a function 
\[
R': \mathcal{A} \times \mathcal{A}^* \rightarrow \mathcal{A}^*\,,
\]
where $R'(c, s, S)$ is the string that results from replacing the $i$th occurrence of character $c$ in $S$ with the $i$ character in $s$. The sequential replacement function will be used by the residual string. Define the set of codebooks $C$ as a sequence of pairs, relating a single character to a string:
\[
\mathcal{C}_{\mathcal{A}} = (\mathcal{A} \times \mathcal{A}^*)^*\,,
\]
and let the cost function be overloaded so that $L(C)$
is the sum of the cost of each entry in $C$. Note that characters on the left of one entry are allowed to appear in the strings on the right of some other entry. Let $f(C)$ define the set of characters that appear as the first element of some entry in $C$. The function~$f$ returns the equivalent, at the character level, of the words that appear in a lexicon or dictionary.

Define an encoding for a given text $S \in \mathcal{A}^*$ as a triple $(C, I, X)$ such that the following hold:
\begin{enumerate}
    \item $C \in \mathcal{C}_{\mathcal{\bar{A}}}$ for some $\mathcal{\bar{A}} \supset \mathcal{A}$, that is, $C$ is a codebook for the alphabet $\mathcal{A}$ that is also allowed contain extra symbols not in $\mathcal{A}$;
    \item $I \in (f(C) \cup \{x\})^*, x \notin \mathcal{\bar{A}}$, that is, $I$ is a string composed of characters that appear as entries in the codebook and a special character $x$ that does not appear in the codebook or in $S$;
    \item $X \in \mathcal{A}^*$;
    \item $R(c_1, s_1, \cdot) \circ \dots \circ R(c_n, s_n, \cdot) \circ R'(x, X, \dot)(I) = S$, that is, if we begin with the string $I$, then replace each occurrence of an entry with the corresponding string in the codebook, followed by replacing all $x$ characters with the substring $X$, the final result is the input string $S$ (this notation uses function currying, where we regard each $R(c,s,\dot)$ as a function from strings to strings, and then compose).
\end{enumerate}
An encoding of $S$ is a choice of codebook, index string and residual string, such that $S$ can be generated by replacing the characters in the index string with the corresponding entries in the codebook, and then replacing the special characters $x$ with the residual string.

Let $\mathcal{E}(S)$ be the set of all such encodings for a given string $S \in \mathcal{A}^*$. The \ac{lcc} score is then defined as follows:
\begin{gather*}
    LCC(S) = L(C) + L(I)\,, \\
    \text{ such that } \exists X \in \mathcal{A}^* \text{ such that }\\
    (C, I, X) = \argmin_{(C', I', X') \in \mathcal{E}(S)} L(C') + L(I') + L(X')\,.
\end{gather*}

\subsection{Formal Description for Continuous Data} \label{app:continuous-full-description}
In the continuous case, we extend the notion of locality, in compositional locality, to also mean local in the space of possible data points: small changes in value should give only small changes in structural representations. For example, in an image, permuting the pixels in a very small patch or changing the RGB composition of one or two pixels in the patch should make little difference to the overall representation. To capture this side of locality, we employ a representation based on clustering, similar to the example used in Section \ref{sec:two-part-descriptions}. In particular, 
For continuous data we cluster progressively larger localised regions, e.g. image patches, using a multivariate \ac{gmm}. The representation model then consists of a sequence of \acp{gmm}, specified by the means and covariance matrices of each cluster. The index consists of an $n$-dimensional 
jagged 
tensor (for images $n=2$, and for audio $n=1$), where each location is either an index to a cluster of the \ac{gmm} or the special symbol $x$. 
By `jagged', we mean that some positions may be missing from the tensor, e.g. the middle row in a three-dimensional jagged matrix may consist of just (1,0) and (1,2). 
The residual consists of the encoding of each point in the input, with respect to its assigned cluster in the \ac{gmm},
under the arithmetic-coding-optimal encoding, 
plus the direct specification, e.g. with 32-bit floats, of each location containing an $x$. 
The data is specified by taking each index appearing in the input tensor, looking up the mean and covariance of the indexed distribution in the \ac{gmm}, then taking the 
arithmetic coding 
code from the corresponding point in the residual tensor, and computing the point it encodes under the indexed distribution. This point is then inserted into the input tensor. 
(The missing locations in the jagged input tensor are chosen so that inserting all indexed points in this way produces a non-jagged tensor as output.) 
One such pass through the input tensor is made for each \ac{gmm} in the sequence, which means that the points taken from one cluster distribution can then be treated themselves as cluster indices, this time into the next \ac{gmm} in the sequence. 
The length of this sequence can be determined automatically as the number above which no new clusters are added to the model. 
This possibility, for points from one cluster`s distribution to refer to indexes into a different clustering model, is what gives this model its compositional structure.

Let $X \subset \mathbb{R}^m, X = x_1,\dots, x_n$ be the input data. Let $c$ be the numerical precision, e.g. $c=32$ in the case of representing real numbers with 32-bit floats. Let $p(x;\mu, \Sigma)$ be the multivariate normal probability of data point $x \in \mathbb{R}^m$ given cluster centroid $\mu \in \mathbb{R}^m$ and diagonal covariance matrix $\Sigma \in \mathbb{R}^m$ (we consider only diagonal covariances to speed up search). Let $g$ be the function that takes as input a partition function $f:\mathbb{R}^m \rightarrow \{0, \dots, n-1\}$, and a data point $x \in X$, and returns the centroid of $x$ under partition $f$. That is $g(f, x) = \frac{1}{|C|} \sum_{y \in C} y$, where $C = \{y \in X | f(y)=f(x)\}$. Let $h$ be the analogous function that returns the diagonal of the covariance matrix of the cluster of point $x$ under $f$, and let $q(x,f) = p(x; g(x, f), h(x,f))$. Let $l(i, f)$ give the number of points assigned to the $i$th cluster under the partition $f$. Then the fit clustering model is given by
\begin{gather*} \label{eq:optimal-partition}
    f^* = \argmin_{f:\mathbb{R}^m \rightarrow \{1, \dots, n\}} \sum_{i=0}^n \min(cm, -\log{q(x_i, f)}) \\
    + \sum_{i=1}^n \log{\frac{n}{l(f(i), f)}}\mathbbm{1}(-\log{q(x_i, f)} < cm)\,,
\end{gather*}
where the last term uses the indicator function $\mathbbm{1}$ to select only those points whose cluster-based description cost is less than their cluster-independent description cost, and the sum represents the Shannon information content of the cluster labels of those points.

\FloatBarrier

\section{Algorithm for the Prime Modulo Text Model} \label{app:bad-model}
Here we describe a model that is very far from being locally compositional, and so ends up giving short descriptions to very unnatural data. The model consists of a list of prime numbers $p_1, \dots, p_k$, and the indexes consist of a sequence of integers $z_1, \dots, z_m$. The data $x_1, \dots, x_n$ was defined by 
\begin{gather} \label{eq:bad-model}
    N = z_1 \nonumber \\
    x_i = chr(\sum_{j=2}^m \frac{(i-z_j)}{p_{l_j}} \bmod N),  l_j \equiv j \;\bmod\; k\; \forall i\,,
\end{gather}
where $chr(i)$ returns the $i$th character in the Roman alphabet. This model takes the first input term $N$ as the length of the output, initializes a string of $N$ zeros, then, for every successive term $z$ in the input, it selects the corresponding prime number, modulo $k$, in the list and increments all elements in the string by their index when the sequence is arranged in powers of $p$ modulo $N$, offset by $z$. The action of this algorithm is shown in Python code below.

\begin{lstlisting}[
    style = myListingStyle,
    ]
def generate_from_inp(inp):
    N = inp[0]
    x = np.zeros(N)
    for p_idx, offset in enumerate(inp[1:]):
        p = primes[p_idx]
        cur_pow = 1
        to_add = []
        for i in range(N):
            to_add.append(cur_pow)
            cur_pow = cur_pow*p \% N
    
        to_add = to_add[-offset:] + to_add[:-offset]
        x += np.array(to_add)
    
    chars = 'abcdefghijklmnopqrstuvwxyzABCDEFGHIJKLMNOPQRSTUVWXYZ '
    return ''.join(chars[int(i)] for i in x\%53)
\end{lstlisting}

Despite obeying a single short equation and being implementable by a short algorithm, this model does not conform to data that is readable to humans. The following is an example of how it can exactly encode a long and ostensibly random string with a relatively short model length

\begin{quote}
    Primes: 2,3,19,5,11 \\
    Input seq: 250,120,24,82,10,15,202 \\
Output seq: qKXVSyiLIyYKyvjXMvZqevkJVy OLjPyBKpccEeQTvFMYyuRzyGBEKKuuTWhkvLdaylFRNMEHKBMxEqknvCgPycIFySsKvsXApTnqKXVSyiLIyYKyvjo pZqOOHnPYKUlWZDwlHaKEQBawWeuGZLCTKuFxWgEBBHyIWkDSKR tKlkfHmNyQydqHFxAvIUNKrtcWPWvQTmRWhGMZccZKxnoHVBdQKgOHnPYKUlWZDwlHaKEQBawWeuGZLTn
\end{quote}
A significant part of the reason this model produces a compact representation for strings that humans do not perceive as regular is that the inclusion of sequences of prime powers modulo $N$ produces highly non-local patterns. 

This example also shows that unnatural strings like the above output can in fact have low Kolmogorov complexity, far lower than the average across strings of a given length, which is known to be very close to the length itself \cite{shen2015around}. In addition to not being able to handle random strings in the manner we desire, we thus see that Kolmogorov complexity is too general to identify the type of structure found in human-readable communicative signals. This generality is also related to its uncomputability \cite{grunwald2007minimum}. It looks for any pattern recognizable by any algorithm, whereas instead we should restrict to a certain form of pattern, where this form is specified by the model. The choice of an appropriate model is therefore essential, and should aim to capture the type of structure that is recognized easily and naturally by humans. 

\FloatBarrier

\subsection{Algorithms to Compute \ac{lcc} score} \label{app:algs}
For strings, the algorithm is as follows. For every substring up to some length $N$ that appears more than once, calculate the change in overall cost from putting it in the codebook and replacing all occurrences in the input string with a new previously unused character that acts as an index to this codebook entry, an action we refer to as `aliasing'. If aliasing reduces the overall cost, then perform it, and move on to the next substring. Repeat until no new substrings are aliased. At this point, the input string will be composed partially of its original characters, and partially of new indexing characters. The final step is to transfer all remaining original characters to the residual string and replace them with $x$ in the input string. 

For continuous data, which we model with the recursive clustering procedure described in Section \ref{subsec:continuous-data}, the algorithm is as follows. Define a sequence of increasing neighbourhood sizes $p_1, \dots, p_m$ (these can be two-dimensional patch sizes in images, or one-dimensional intervals in time series data). For $p=p_1, \dots, p_m$, fit a \ac{gmm} on all neighbourhoods of size $p$ in the input. (For audio, because time series data is one dimensional, the neighbourhood sizes, $p_1, \dots, p_m$, increase along the time axis only.) Compute the bitcost of representing the data with this fit \ac{gmm}, which is equal to the number of bits to represent the components of the \ac{gmm} plus the number for the description of each point in the input, where the latter is either the cost of the cluster index plus the arithmetic coding residual, or the direct cost with e.g. 32-bit floats, whichever is smaller. (The bit precision for these floats need not be set arbitrarily to 32, but can be computed from the data, as the smallest precision needed to represent every distinct value that appears in the input.) We search for the optimal number of components in the \ac{gmm} by computing all up to some fixed value $K$, and selecting that with the overall lowest bitcost. For our experiments, we use $K=15$. Then, replace each neighbourhood in the input with its assigned cluster index, or $x$ if it was selected to be represented directly, and repeat for the next neighbourhood size. Continue repeating until the optimal representation contains no clusters and all $x$`s, or until some fixed threshold is reached, $m$ in this example. In the experiments below, we use a threshold of 4. As before, the complexity score is the sum of the description lengths for the model and the input, which here consist of a sequence of \acp{gmm} and a jagged tensor, respectively. 

\section{Wikipedia Articles in Language Text Experiments} \label{app:wikis-list}
The following are the titles of all the Wikipedia articles used in the experiments on text complexity from Section \ref{sec:discrete-data}, Figure \ref{fig:nl-text-results}.

\begin{itemize}
    \item Life
    \item Computation
    \item Architecture
    \item Water
    \item Plants
    \item Aurora Borealis
    \item Chemistry
    \item Animals
    \item Trees
    \item Ocean
    \item Music
\end{itemize}

\section{Full Numeric Results} \label{app:full-results}

\FloatBarrier

\begin{table}[]
    \centering
\begin{tabular}{lllll}
\toprule
 & Model Cost & Idx Cost & LCCScore & Residual Cost \\
\midrule
text-en & 1442.38 (58.03) & 13195.02 (225.99) & 14637.40 (195.98) & 9393.79 (457.05) \\
text-de & 1375.24 (46.37) & 13342.95 (122.85) & 14718.19 (101.97) & 9653.65 (168.47) \\
text-ie & 1756.35 (142.88) & 11756.63 (335.73) & 13512.98 (220.23) & 8647.10 (219.83) \\
simp-en & 20.26 (0.00) & 5801.94 (0.70) & 5822.21 (0.70) & 0.00 (0.00) \\
rand & 0.00 (0.00) & 0.00 (0.00) & 0.00 (0.00) & 46801.71 (3.60) \\
repeat2 & 2.00 (0.00) & 0.00 (0.00) & 2.00 (0.00) & 0.00 (0.00) \\
repeat5 & 11.21 (0.27) & 0.00 (0.00) & 11.21 (0.27) & 0.00 (0.00) \\
repeat10 & 29.13 (0.52) & 0.00 (0.00) & 29.13 (0.52) & 0.00 (0.00) \\
\bottomrule
\end{tabular}

    \caption{Full, raw results for text with std. dev. in parentheses.}
    \label{tab:text-results}
\end{table}

\begin{table}[]
\resizebox{\textwidth}{!}{
    \centering
\begin{tabular}{lllll}
\toprule
 & LCCScore & Residual Cost & Model Cost & Idx Cost \\
\midrule
bell & 2043.14 (1110.04) & 193898.73 (4841.73) & 782.77 (312.90) & 1260.38 (797.13) \\
white-noise & 273.41 (0.38) & 191116.32 (58.84) & 273.41 (0.38) & 0.00 (0.00) \\
walla & 7036.59 (190.64) & 175630.20 (3388.02) & 2270.64 (67.02) & 4765.94 (132.75) \\
tuning-fork & 4808.17 (1001.92) & 178929.18 (4271.23) & 1609.30 (289.55) & 3198.87 (717.18) \\
irish-speaker-1 & 9387.16 (222.87) & 166074.92 (2763.26) & 2786.92 (70.64) & 6600.24 (190.05) \\
irish-speaker-2 & 9674.27 (234.00) & 167794.02 (2065.28) & 2778.14 (78.03) & 6896.13 (165.98) \\
english-speaker-1 & 9505.16 (208.75) & 170094.87 (3148.59) & 2802.31 (49.05) & 6702.85 (179.56) \\
english-speaker-2 & 9617.23 (323.11) & 165068.49 (1455.42) & 2862.03 (96.06) & 6755.19 (244.95) \\
german-speaker-1 & 9525.03 (441.60) & 160626.07 (1362.32) & 2802.65 (127.31) & 6722.38 (316.96) \\
german-speaker-2 & 9703.21 (198.38) & 167156.56 (1049.71) & 2819.16 (35.16) & 6884.06 (174.63) \\
rain & 294.91 (0.43) & 200476.58 (43.64) & 294.91 (0.43) & 0.00 (0.00) \\
\bottomrule
\end{tabular}

}
    \caption{Full, raw results for audio with std. dev. in parentheses.}
    \label{tab:full-audio-results}
\end{table}

\begin{table}[]
    \centering
    \resizebox{\textwidth}{!}{
\begin{tabular}{lllll}
\toprule
 & Model Cost & Idx Cost & LCCScore & Residual Cost \\
\midrule
imagenet & 6865.69 (138.88) & 492500.82 (13029.60) & 499366.51 (13159.53) & 10161027.93 (214368.74) \\
cifar10 & 7114.44 (62.83) & 507249.40 (4214.13) & 514363.84 (4239.33) & 10154058.99 (54263.77) \\
stripes & 4800.59 (27.60) & 242604.22 (3057.78) & 247404.82 (3063.10) & 8175600.59 (11705.87) \\
halves & 4838.60 (10.60) & 181564.76 (482.96) & 186403.35 (485.99) & 7964144.40 (1590.83) \\
rand & 384.00 (0.00) & 0.00 (0.00) & 384.00 (0.00) & 14427771.33 (49.71) \\
\bottomrule
\end{tabular}
}
\caption{Full, raw results for images with std. dev. in parentheses.}
    \label{tab:image-results}
\end{table}

\FloatBarrier
\section{Derivation of Randomness Threshold} \label{app:rand-thresh-deriv}
In Section \ref{sec:discussion}, \eqref{eq:str-rand-threshold} showed the threshold above which our framework will alias a given substring. This threshold arises directly from the imperative to select the shortest description. Suppose we are given a string $S$ and are considering whether to alias some substring $s$. Let $C_1$ be the cost of the non-aliased portion of $S$, that is, the subsequence with all occurrences of $s$ removed. Let $C_2$ be the cost of the same substring, assuming we have aliased $s$. The total cost, if we do not alias $s$, is then $C_1 + nl_1$, where $l_1$ is the cost of describing $s$. The total cost if we do alias is $C_2 + nl_2$, where $l_2$ is the cost of describing $s$ under the new encoding after aliasing. Aliasing changes the distribution of characters, and hence the optimal encoding, however if we assume this change is small for the string $s$ itself, then $l_1 \approx l_2$, which we write as $l$. Then, we alias if and only if

\begin{gather}
    C_1 + nl > l + C_2 \\
    (n-1)l > C_2 - C_1 \\
    n > 1 + \frac{C_2 - C_1}{l} \,.
\end{gather}
Now, let $c_i$ be the number of times the $i$th character appears in the non-aliased portion of $S$, and let $f_i$ and $f_i'$ be, respectively, the frequency of the $i$th character across the whole string before and after aliasing. Then 
\begin{align*}
c_2 - C_1 =& (- \sum_i c_i \log{f_i}) - (- \sum_i c_i \log{f_i'}) \\
          =& \sum_i c_i \log{f_i'} - c_i \log{f_i} \\
          =& \sum_i c_i \log{\frac{f_i'}{f_i}}\,,
\end{align*}
which gives the bound in \eqref{eq:str-rand-threshold}. Note that $i$ ranges over the sequence of characters, so those that appear multiple times are counted multiple times. This also means the threshold increases as the length of the input data increases. 

\end{document}